%% file: main.tex
\documentclass[11pt, a4paper]{lumia}

\usepackage[sort&compress]{natbib}
\input{notations}

\usepackage{graphicx}           
\usepackage{tikz}               
\usepackage[edges]{forest}      

\usepackage{url}                
\usepackage{xurl}               

\usepackage{array}              
\usepackage{longtable}          
\usepackage{multirow}           
\usepackage{makecell}           
\usepackage{ragged2e}           

\usepackage{mathtools}          
\usepackage{nicefrac}           

\usepackage{algorithm}          
\usepackage{algorithmicx}       
\usepackage{algpseudocode}      
\usepackage{listings}           

\usepackage{subcaption}         
\usepackage{wrapfig}            
\usepackage[export]{adjustbox}  

\usepackage[commandnameprefix=always]{changes}
\usepackage{xspace}             
\usepackage[normalem]{ulem}     
\usepackage{CJKutf8}            

\usepackage[tikz]{bclogo}       
\usepackage[framemethod=tikz]{mdframed} 

\usepackage{lipsum}             
\usepackage{tocloft}            
\usepackage{afterpage}          
\usepackage{bbding}             
\usepackage{epigraph}           
\usepackage{minitoc}            
\usepackage{multicol}           
\usepackage{textgreek}          
\usepackage{wrapfig}

\newcolumntype{P}[1]{>{\RaggedRight\arraybackslash}p{#1}}

\definecolor{uclablue}{RGB}{39, 116, 174}
\definecolor{bigaired}{RGB}{156, 0, 0}
\definecolor{myblue}{HTML}{598BE7}
\definecolor{mildblue}{RGB}{31,119,180}
\definecolor{sectionblue}{RGB}{70, 130, 180}
\definecolor{methodblue}{RGB}{0, 150, 136}
\definecolor{bgblue}{RGB}{245,243,253}
\definecolor{ttblue}{RGB}{91,194,224}
\definecolor{mygreen}{rgb}{0.64, 0.56, 0.88}
\definecolor{myyellow}{rgb}{0.68, 0.6, 0.1}
\definecolor{fancygreen}{rgb}{0.33, 0.68, 0.20}
\definecolor{salmon}{rgb}{0.94, 0.52, 0.49}
\definecolor{tablegreen}{rgb}{0.82, 0.94, 0.75}
\definecolor{tableblue}{rgb}{0.81, 0.90, 0.94}
\definecolor{tablered}{rgb}{0.97, 0.85, 0.85}
\definecolor{tableorange}{rgb}{0.96, 0.85, 0.81}
\definecolor{myorange}{rgb}{1.0, 0.49, 0.0}
\definecolor{tlgreen}{rgb}{0.33, 0.68, 0.20}
\definecolor{darkgreen}{RGB}{0,100,0}
\definecolor{darkred}{RGB}{200, 0, 0}
\definecolor{customyellow}{HTML}{FFFACD}
\definecolor{refinegreen}{RGB}{0, 128, 75}
\definecolor{scoregreen}{RGB}{34, 139, 34}
\definecolor{hidden-blue}{RGB}{194,232,247}
\definecolor{hidden-black}{RGB}{20,68,106}
\definecolor{yes}{HTML}{C6EFCE}
\definecolor{no}{HTML}{FFC7CE}
\definecolor{partial}{HTML}{FFEB9C}
\definecolor{external}{HTML}{D9E1F2}
\definecolor{hdr}{HTML}{F2F2F2}
\definecolor{GRPOrow}{gray}{0.96}
\definecolor{FlowRLrow}{RGB}{225,236,255}
\definecolor{FlowBlue}{RGB}{80,120,210}
\definecolor{GRPOGray}{gray}{0.35}

\hypersetup{
    colorlinks=true, 
    citecolor=uclablue, 
    linkcolor=bigaired,
    urlcolor=darkblue
}

\setlist[itemize]{leftmargin=20pt, noitemsep, topsep=0pt}

\NewDocumentCommand{\kaiyan}{mO{}}{\textcolor{purple}{\textsuperscript{\textit{kaiyan}}\textsf{\textbf{\small[#1]}}}}
\NewDocumentCommand{\yuxin}{mO{}}{\textcolor{cyan}{\textsuperscript{\textit{yuxin}}\textsf{\textbf{\small[#1]}}}}
\NewDocumentCommand{\bx}{mO{}}{\textcolor{green}{\textsuperscript{\textit{bx}}\textsf{\textbf{\small[#1]}}}}
\NewDocumentCommand{\at}{mO{}}{\textcolor{red}{\textsuperscript{\textit{AT}}\textsf{\textbf{\small[#1]}}}}
\NewDocumentCommand{\re}{mO{}}{\textcolor{blue}{\textsuperscript{\textit{RE}}\textsf{\textbf{\small[#1]}}}}
\NewDocumentCommand{\ybsun}{mO{}}{\textcolor{magenta}{\textsuperscript{\textit{youbang}}\textsf{\textbf{\small[#1]}}}}
\NewDocumentCommand{\runze}{mO{}}{\textcolor{orange}{\textsuperscript{\textit{runze}}\textsf{\textbf{\small[#1]}}}}
\NewDocumentCommand{\add}{mO{}}{\textcolor{darkgreen}{\textsuperscript{\textit{Maybe Consider Discuss}}\textsf{\textbf{[#1]}}}}

\newcommand{\cmark}{\textcolor{darkgreen}{\boldmath$\checkmark$}}
\newcommand{\xmark}{\textcolor{darkred}{\boldmath$\times$}}

\newenvironment{itemize*}%
 {\leftmargini=10pt\begin{itemize}%
  \setlength{\itemsep}{0pt}%
  \setlength{\parskip}{0pt}%
  }%
 {\end{itemize}}

\newenvironment{enumerate*}%
 {\begin{enumerate}%
  \setlength{\itemsep}{0pt}%
  \setlength{\parskip}{0pt}}%
 {\end{enumerate}}

\newcommand{\cellstatus}[1]{%
  \begingroup
  \StrTrim{#1}[\statusval]%
  \IfStrEq{\statusval}{Yes}{\cellcolor{yes}\cmark}{}%
  \IfStrEq{\statusval}{No}{\cellcolor{no}\xmark}{}%
  \IfBeginWith{\statusval}{Yes (}{\cellcolor{yes}\cmark~\textit{\statusval\unskip}}{}%
  \IfStrEq{\statusval}{Partial}{\cellcolor{partial}\textbf{Partial}}{}%
  \IfStrEq{\statusval}{External}{\cellcolor{external}\textbf{External}}{}%
  \endgroup
}

\newtcolorbox{myboxi}[1][]{
  breakable,
  title=#1,
  colback=red!5,
  colbacktitle=red!5,
  coltitle=black,
  fonttitle=\bfseries,
  bottomrule=0pt,
  toprule=0pt,
  leftrule=2pt,
  rightrule=2pt,
  titlerule=0pt,
  arc=0pt,
  outer arc=0pt,
  colframe=red,
}

\newtcolorbox{myboxnote}[1][]{
  breakable,
  title=#1,
  colback=orange!0,
  colbacktitle=orange!0,
  coltitle=black,
  fonttitle=\bfseries,
  bottomrule=0pt,
  toprule=0pt,
  leftrule=2pt,
  rightrule=2pt,
  titlerule=0pt,
  arc=0pt,
  outer arc=0pt,
  colframe=orange,
}

\newtcolorbox{myboxii}[1][]{
  breakable,
  freelance,
  title=#1,
  colback=white,
  colbacktitle=white,
  coltitle=black,
  fonttitle=\bfseries,
  bottomrule=0pt,
  boxrule=0pt,
  colframe=white,
  overlay unbroken and first={
  \draw[red!75!black,line width=3pt]
    ([xshift=5pt]frame.north west) -- 
    (frame.north west) -- 
    (frame.south west);
  \draw[red!75!black,line width=3pt]
    ([xshift=-5pt]frame.north east) -- 
    (frame.north east) -- 
    (frame.south east);
  },
  overlay unbroken app={
  \draw[red!75!black,line width=3pt,line cap=rect]
    (frame.south west) -- 
    ([xshift=5pt]frame.south west);
  \draw[red!75!black,line width=3pt,line cap=rect]
    (frame.south east) -- 
    ([xshift=-5pt]frame.south east);
  },
  overlay middle and last={
  \draw[red!75!black,line width=3pt]
    (frame.north west) -- 
    (frame.south west);
  \draw[red!75!black,line width=3pt]
    (frame.north east) -- 
    (frame.south east);
  },
  overlay last app={
  \draw[red!75!black,line width=3pt,line cap=rect]
    (frame.south west) --
    ([xshift=5pt]frame.south west);
  \draw[red!75!black,line width=3pt,line cap=rect]
    (frame.south east) --
    ([xshift=-5pt]frame.south east);
  },
}

\tcbset{
  takeawaysbox/.style={
    title=Takeaways,
    colback=lightblue!80,
    colframe=black,
    fonttitle=\bfseries\small,
    coltitle=white,
    colbacktitle=black,
    enhanced,
    attach boxed title to top left={xshift=2.5mm,yshift=-2.5mm},
    boxed title style={rounded corners, size=small, colframe=black, colback=black},
    width=\linewidth,
    arc=3.5mm
  }
}

\mdfdefinestyle{mystyle}{%
  rightline=true,
  innerleftmargin=10,
  innerrightmargin=10,
  outerlinewidth=3pt,
  topline=false,
  rightline=true,
  bottomline=false,
  skipabove=\topsep,
  skipbelow=\topsep
}

\tikzset{%
    every node/.style={font=\tiny},
    parent/.style =          {align=center,text width=2cm,rounded corners=3pt, line width=0.3mm, fill=gray!10,draw=gray!80},
    child/.style =           {align=center,text width=2.0cm,rounded corners=3pt, fill=blue!10,draw=blue!80,line width=0.3mm},
    grandchild/.style =      {align=center,text width=2cm,rounded corners=3pt},
    greatgrandchild/.style = {align=center,text width=1.5cm,rounded corners=3pt},
    greatgrandchild2/.style = {align=center,text width=1.5cm,rounded corners=3pt},    
    referenceblock/.style =  {align=center,text width=1.5cm,rounded corners=2pt},
    pretrain/.style =           {align=center,text width=2.0cm,rounded corners=3pt, fill=blue!10,draw=blue!80,line width=0.3mm},   
    pretrain_work/.style =           {align=center, text width=8.5cm,rounded corners=3pt, fill=blue!10,draw=blue!0,line width=0.3mm},  
    template/.style =           {align=center,text width=2.0cm,rounded corners=3pt, fill=red!10,draw=red!80,line width=0.3mm},   
    template_work/.style =           {align=center,text width=8.5cm,rounded corners=3pt, fill=red!10,draw=red!0,line width=0.3mm},    
    answer/.style =           {align=center,text width=2.0cm,rounded corners=3pt, fill= cyan!10,draw= cyan!80,line width=0.3mm},   
    answer_work/.style =           {align=center,text width=8.5cm,rounded corners=3pt, fill= cyan!10,draw= cyan!0,line width=0.3mm},      
    multiple/.style =           {align=center,text width=2.0cm,rounded corners=3pt, fill= orange!10,draw= orange!80,line width=0.3mm},   
    multiple_work/.style =           {align=center,text width=8.5cm,rounded corners=3pt, fill= orange!10,draw= orange!0,line width=0.3mm},        
    tuning/.style =           {align=center,text width=2.0cm,rounded corners=3pt, fill= magenta!10,draw= magenta!80,line width=0.3mm},   
    tuning_work/.style =           {align=center,text width=8.5cm,rounded corners=3pt, fill= magenta!10,draw= magenta!0,line width=0.3mm},          
}

\newcommand{\lstbg}[3][0pt]{{\fboxsep#1\colorbox{#2}{\strut #3}}}

\lstdefinelanguage{diff}{
  basicstyle=\ttfamily\small,
  morecomment=[f][\lstbg{red!20}]-,
  morecomment=[f][\lstbg{green!20}]+,
}

\lstdefinelanguage{diffpython}{
  language=diff,
  morekeywords={def, if, else, for, while, return, import, from, as, class, with, try, except, finally, raise, lambda, and, or, not, in, is, None, True, False},
  morecomment=[l]{\#},
  morestring=[b]",
  morestring=[b]',
}

\setheadertext{LUMIA Lab}

\correspondingemail{\emailicon lihe27341@gmail.com \emailicon lin.zhouhan@gmail.com \quad $^*$ Equal Contribution. $^\ddagger$ Corresponding Author. }

\renewcommand{\today}{2026-03-02}

\title{PonderLM-3: Adaptive Token-Wise Pondering with Differentiable Masking}
\setheadertitle{PonderLM-3: Adaptive Token-Wise Pondering with Differentiable Masking}

\author{%
  He Li$^{1*}$, Feichen Song$^{1*}$, Boyi Zeng$^{1}$, Shixiang Song$^{1,2}$, Zhiqin John Xu$^{3}$, Ziwei He$^{2}$, Zhouhan Lin$^{1,2,4\ddagger}$\\
  $^1$ LUMIA Lab, School of Artificial Intelligence, Shanghai Jiao Tong University \\
  $^2$ Shanghai AI Laboratory, $^3$ Shanghai Jiao Tong University, $^4$ Shanghai Innovation Institute
}

\begin{document}

\begin{abstract}
Test-time scaling has shown that allocating more additional computation at inference can improve generation quality, motivating a natural follow-up question: where should this computation be spent? Building on this insight, we introduce \textit{PonderLM-3}, a pretraining framework for token-wise adaptive pondering that learns to selectively allocate additional computation under purely self-supervised objectives, built on top of the PonderLM-2 backbone. This makes additional inference computation an allocatable per-token resource, so tokens receive more computation only when it is beneficial, rather than paying a uniform extra cost. To make this allocation learnable while maintaining train–inference consistency, PonderLM-3 injects a differentiable attention mask during pretraining and pairs it with a matching hard pruning rule at inference. PonderLM-3 defines a stronger Pareto frontier: compared with existing recursive or adaptive baselines, it achieves lower pretraining perplexity at equal inference FLOPs. On downstream benchmarks, PonderLM-3 attains comparable performance to fixed-step PonderLM-2 under the same maximum number of additional computation steps, while using fewer inference FLOPs in practice. Overall, PonderLM-3 provides an end-to-end differentiable and train–inference consistent framework for token-wise adaptive computation, enabling additional inference compute to be allocated where it is most useful rather than paid uniformly by every token.
\end{abstract}

\maketitle


\begin{figure*}[!htbp]
    \centering
    \includegraphics[width=\textwidth]{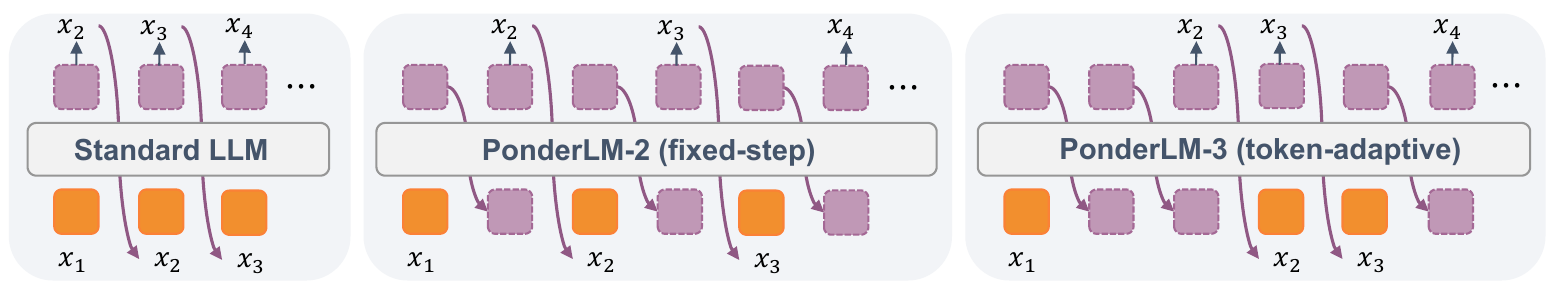}
    \caption{From fixed compute to token-wise allocation. Standard LMs decode with one forward pass per token. PonderLM-2 applies fixed pondering steps, turning extra compute into a uniform per-token tax. PonderLM-3 makes refinement depth token-dependent, allocating additional internal updates only when they provide non-trivial marginal gains.}
    \label{fig:paradigm-compare}

    \vskip -0.1in
\end{figure*}

\section{Introduction}
Recent progress in test-time scaling~\cite{snell2024scaling,geiping2025scaling} shows that, under a fixed parameter budget, the allocation of additional inference-time compute can improve the quality of generation on challenging reasoning and decision-making tasks, offering a complementary axis beyond the size of the model. Building on this insight, a growing line of work aims to internalize such extra compute into self-supervised pretraining, such as LoopedLM~\cite{giannou2023looped,saunshi2025reasoning} and PonderLMs~\cite{zeng2025pretraining1,zeng2025pretraining2}. By incorporating iterative additional computation steps directly into the language modeling objective, these methods learn multi-step ``thinking'' from large-scale unlabeled corpora, without requiring human-annotated supervision.

However, existing LoopedLM and PonderLMs typically use a fixed number of computation steps per token: inference cost grows linearly with the step budget, turning extra compute into an unavoidable ``fixed tax''. This uniform policy clashes with the fact that compute demand in language generation is highly non-uniform--many tokens are merely local continuation or copying, while a smaller fraction disproportionately affects semantic trajectory and error propagation. Fixed-step computation thus wastes compute on easy tokens, may trigger overthinking~\cite{fu2025think} that degrades predictions, and still under-allocate compute to the few hard tokens that benefit most from additional computation steps. Therefore, We aim for models that not only ``think longer'', but also assign an appropriate thinking depth per token and stop when marginal gains diminish---adapting extra compute to the tokens that truly need it without sacrificing generation fidelity.

This objective reframes inference compute from a fixed overhead into an allocatable resource: the model should decide how far to refine and when to stop during generation. From this perspective, Adaptive Computation Time (ACT)--style dynamic halting provides a natural framework. Despite progress, existing ACT paradigms still commonly exhibit three limitations: (i) limited adaptivity, where compute-saving policies such as pruning ratios, token budgets, or halting priors are fixed by design rather than learned from data-driven token difficulty~\cite{raposo2024mixture,bae2025mixture,chen2025inner,hwang2025dynamic,banino2021pondernet}; (ii) non-end-to-end training under self-supervised pretraining, as many methods rely on SFT/RL, extra supervision, multi-stage procedures, or post-hoc threshold calibration~\cite{fu2025think,he2025learning,jolicoeur2025less,fu2025deep,schuster2022confident,scalena2025eager}; (iii) structural overhead during acceleration, such as draft-and-verify, early-exit rollback, or piecewise static policy compositions, whose extra costs can trigger more frequently on hard tokens, sometimes wasting compute where deeper processing is most needed~\cite{leviathan2023fast,elhoushi2024layerskip}; and
(iv) train–inference mismatch, since halting decisions must be learned under parallel training but executed under sequential generation, making it non-trivial to align skipping behavior during training with its inference-time instantiation. 

In this work, we propose \textbf{PonderLM-3}, a data-driven, token-wise halting mechanism for pondering-style pretraining models, trained end-to-end under self-supervised language modeling. Building on the PonderLM-2~\cite{zeng2025pretraining2} backbone that supports efficient parallel training via Jacobi iterations, we retain training efficiency and train–inference consistency while learning per-token pondering-step allocation. To make inference-time hard stopping learnable under self-supervised pretraining, we let a lightweight router produce a \emph{differentiable attention mask} that modulates the contributions of later pondering steps during training, providing a continuous approximation of step skipping optimized directly from the next-token objective. At inference, this learned gate is instantiated as a hard-stopping criterion, allowing the model to execute only the necessary pondering steps. This approach achieves tangible reductions in inference latency while maintaining generation fidelity, offering a more granular and controllable compute-quality trade-off.

Our contributions can be summarized as follows: We (i) make additional inference computation allocatable at the token level, (ii) learn a train–inference consistent halting mechanism via a differentiable attention mask, and (iii) provide evidence that compute is concentrated on intrinsically hard tokens.

\section{Related Work}
\subsection{Test-time scaling}
Test-time scaling aims to improve output quality by allocating additional computation at inference time without modifying model parameters. Existing approaches can be broadly categorized into \textit{parallel} and \textit{sequential} methods. Parallel methods typically sample multiple candidate outputs for the same input and then aggregate them via voting or scoring to improve robustness and accuracy, such as majority voting~\cite{wang2023guiding} and best-of-$N$ (BoN) selection \cite{cobbe2021training,sun2024fast}. Sequential methods, in contrast, introduce extra intermediate computation and state updates within a single generation process, and can be further divided into \textit{vertical} and \textit{horizontal} forms. Vertical methods perform multiple iterations or recurrent updates for each token or local step, effectively deepening computation at the same generation position\cite{giannou2023looped,saunshi2025reasoning,zeng2025pretraining1,dehghani2018universal}. Horizontal methods explicitly expand the reasoning trajectory by introducing additional reasoning tokens or intermediate steps to guide subsequent generation, such as Chain-of-Thought (CoT)~\cite{wei2022chain}, Coconut~\cite{hao2024training}, Quiet-STaR~\cite{zelikman2024quiet}, and PonderLM-2~\cite{zeng2025pretraining2}. Since this paper focuses on mechanisms for variable compute budgets in pretraining and inference, we emphasize ACT-style dynamic halting methods and their limitations in the following.

\subsection{Adaptive Computation Time}
Adaptive Computation Time \cite{graves2016adaptive} was first formulated by Graves in recurrent networks, enabling the model to dynamically choose the number of internal updates per input position. This idea was later adopted in Transformer-style architectures, notably by the Universal Transformer\cite{dehghani2018universal}, which combines parameter sharing with a dynamic halting mechanism, and by Depth-Adaptive Transformer-style early-exit models that learn or estimate input-dependent computational depth.

\begin{table*}[!t]
\centering
\small
\setlength{\tabcolsep}{4pt} 
\renewcommand{\arraystretch}{1.2} 

\caption{Adaptive-compute methods vs.\ ours.}
\label{tab:act-compare}
\begin{tabularx}{\textwidth}{@{} l l X X @{}}
\toprule
\textbf{Method} & \textbf{Training Paradigm} & \textbf{Adaptivity Source} & \textbf{Constraints / Priors} \\
\midrule
PonderNet & Pretraining & Learned stopping distribution & Geometric prior \\
MoR / MoD / ITT & Pretraining & Router-based token selection & Fixed continuation ratio \\
Think-at-Hard & SFT (LoRA) & Supervised step targets & External step labels \\
Learning to Ponder & RL (GRPO) & Learned controller & \makecell[l]{Frozen backbone; \\ reward-shaped budget} \\
CALM / DAT & Post-hoc & Confidence-based stopping & Calibration of threshold \\
LayerSkip & Pretraining & Draft--verify & Static exit policy \\
\midrule
PonderLM-3 (ours) & Pretraining & \makecell[l]{Router-predicted \\differentiable attention mask} & None \\
\bottomrule
\end{tabularx}
\end{table*}

\paragraph{Fixed-budget and prior-based routing.}
These methods allow token-level dynamic computation, yet the overall budget is often constrained by fixed continuation ratios or hand-specified priors. For instance, MoR~\cite{bae2025mixture} applies a top-\(k\) continuation rule so that only a preset fraction of tokens remains active per iteration ; related approaches include MoD~\cite{raposo2024mixture},ITT~\cite{chen2025inner}.H-Net~\cite{hwang2025dynamic} controls compute through dynamic chunking enforcing a target compression ratio through a specialized ratio-style loss. Furthermore, PonderNet~\cite{banino2021pondernet} shapes halting behavior with a regularizer aligned to a geometric prior. Consequently, the resulting adaptivity remains heavily dictated by these imposed budgets or structural priors, rather than emerging as a purely data-driven response to token difficulty.

\paragraph{Supervision, RL, and calibration-based stopping}
Another line of research derives stopping policies and step allocation with training paradigms beyond end-to-end self-supervised pretraining, including SFT/RL objectives and multi-stage training pipelines~\cite{fu2025think,he2025learning,jolicoeur2025less}, or with post-hoc calibration procedures~\cite{fu2025deep,schuster2022confident,scalena2025eager}. Think-at-Hard, for instance, first runs offline inference to obtain token-level annotations of how many steps to compute, and then trains a routing or halting policy in SFT using these external targets. Alternatively, training-free variants select thresholds on a calibration set based on confidence or similarity to trigger early stopping or additional computation. However, such heavy reliance on auxiliary signals and multi-step procedures may compromise the robustness of the resulting stopping behavior across diverse datasets and deployment environments.

\paragraph{Draft-and-verify decoding}
  Draft-and-verify decoding accelerates generation by proposing candidate tokens with a cheaper draft model and validating them with a more capable verifier \cite{leviathan2023fast,elhoushi2024layerskip}). LayerSkip uses a static exit-depth schedule for drafting and falls back via verification/rollback when needed. Since the drafting policy is not token-difficulty adaptive, hard tokens more often incur failures and rollback, concentrating overhead where deeper computation is most needed.

\section{Method}
\label{sec:method}

\subsection{Background: Jacobi-aligned Latent-Thought Backbone}
\label{sec:background}

We build on the pretraining framework of PonderLM-2~\cite{zeng2025pretraining2}, which inserts \(K\) internal latent pondering steps between adjacent observed tokens, forming a sequence of hidden states \(\{h_t^{(k)}\}_{k=0}^{K-1}\) at each position. At inference time, the model repeatedly feeding the current position’s last-layer hidden state back as the input embedding for subsequent pondering steps, yielding a sequence of additional computation steps at the same token position. At training time, PonderLM-2 adopts a Jacobi-aligned approximation that rewrites the recurrent updates as parallel iterations: in each iteration, a single Transformer forward pass updates all latent positions simultaneously, and a small number of iterations is performed to approach the fixed point that matches inference-time behavior.

\subsection{Token-Wise Adaptive Pondering Modules}
PonderLM-3 consists of three key components: (1) a router that predicts the step distribution and a derived mask score; (2) a differentiable attention masking mechanism parameterized by the mask score; and (3) weighted hidden-state integration across pondering steps.

\subsubsection{Step distribution and mask score}
For each token position $t$, a lightweight router conditions on the step-0 hidden state $h_t^{(0)}$ and predicts a \textit{step distribution} over the number of pondering steps:
\[
s_{t,k} = \Pr(\text{token } t \text{ uses exactly } k \text{ pondering steps}),
\]
\[
k=0,\ldots,K.
\]
From the step distribution, we compute a monotone \textit{mask score} $w_{t,k}$ using the tail Cumulative Distribution Function (tail CDF):
\[
w_{t,k} = \sum_{j=k}^{K} s_{t,j}, \qquad k=0,\ldots,K.
\]
This tail-CDF construction can be viewed as a reverse cumulative sum of the step distribution, and it ensures that $w_{t,k}$ is non-increasing with respect to the step index $k$. Intuitively, $w_{t,k}$ measures how much probability mass remains for continuing beyond step $k$.

Moreover, the mask score plays a shared role in both training and inference: it parameterizes a \emph{differentiable attention mask} that controls the attention weight of later-step latent states. Inference further applies a hard stopping rule based on $w$ to skip the remaining pondering steps when their contributions are numerically negligible.

\begin{figure*}[!t]
    \centering
    \includegraphics[width=\textwidth]{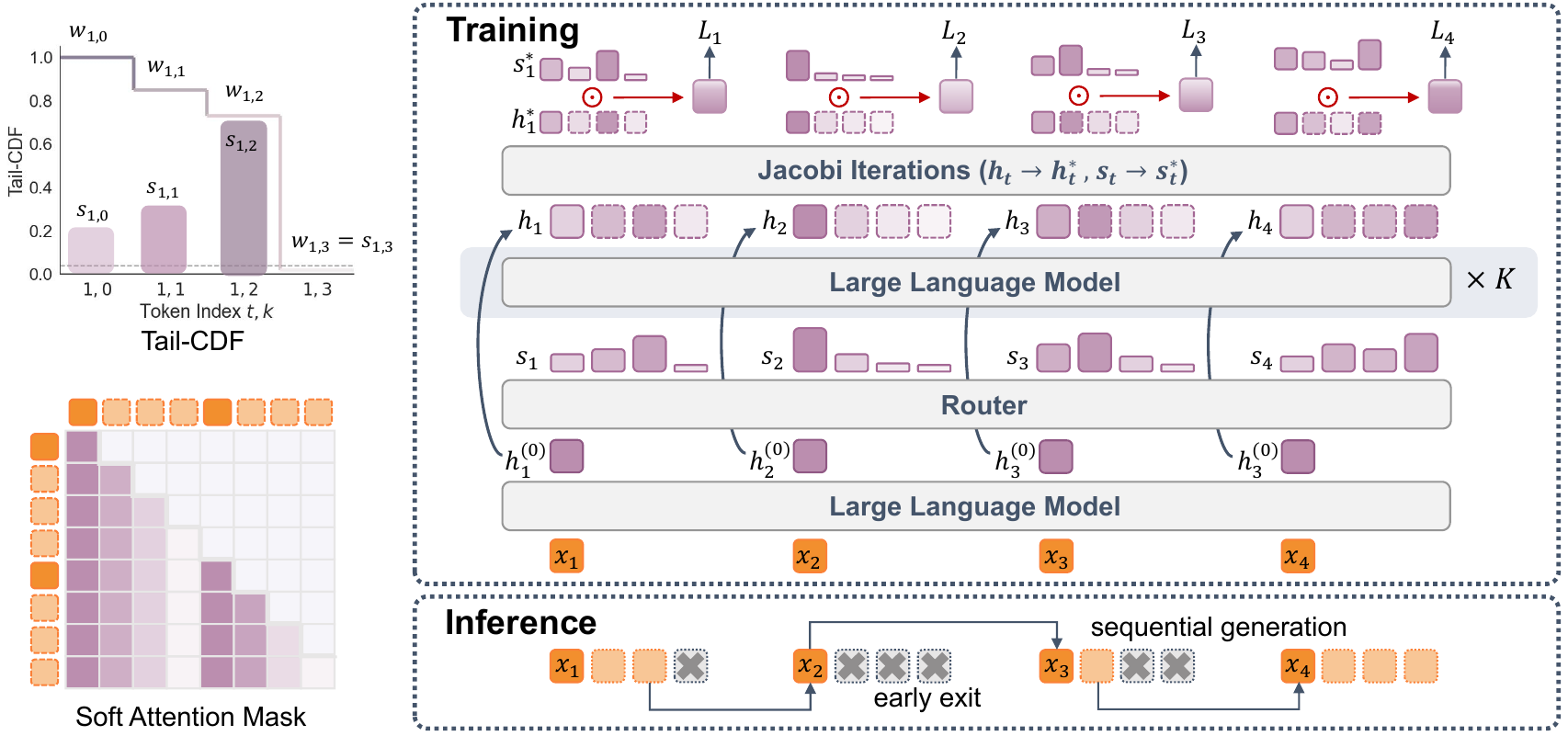}
    \caption{Token-wise adaptive pondering in PonderLM-3.
    \textbf{(1)} A lightweight router predicts the step distribution $s_{t,k}$ from $h_t^{(0)}$, and obtains a monotone mask score $w_{t,k}$ via the tail-CDF of $s$; $w_{t,k}\in[0,1]$ serves as a soft attention mask (color intensity indicates $w$).
    \textbf{(2)} the same LLM performs $K$ pondering steps; at each step, the current hidden state is inserted as a latent token at its corresponding position to form an interleaved sequence, and $w_{t,k}$ is applied in attention as a soft mask. Then we use Jacobi iterations to approximate the sequential inference dynamics, and the output uses the $s_{t,k}$-weighted fused hidden state.
    \textbf{(3)} Inference: token-by-token decoding uses the same attention masking and fusion mechanism, and early exits once $w_{t,k}<\tau$, skipping remaining pondering steps.}
    \label{fig:main}
\end{figure*}

\subsubsection{Differentiable attention mask}
We use the mask score $w$ to construct a differentiable attention mask that smoothly down-weights later-step latent states in attention. Concretely, we inject $\log w$ as an additive bias to attention logits:
\[
\mathrm{Attn}(Q,K,V)=\mathrm{Softmax}\!\left(\frac{QK^\top}{\sqrt d}+M+\log w\right)V,
\]
where $M$ is the standard attention mask (causal and padding). This injection lets $w$ continuously control the visibility of the corresponding latent position: as $w\to 0$, its attention weight smoothly approaches zero, making that position effectively ``invisible'' to attention during training. This provides a soft, end-to-end differentiable mechanism that mirrors inference-time hard stopping: when the remaining probability mass becomes numerically negligible, later steps exert vanishing influence through attention, and can be safely skipped at inference with minimal behavioral discrepancy.

In addition, our implementation is compatible with FlashAttention-2: although standard FA2 kernels primarily support boolean masks, we use an equivalent formulation based on an augmented-matrix construction of the $Q/K/V$ tensors to realize the same effect of the differentiable attention mask while retaining FA2 acceleration; details are provided in the appendix~\ref{appendix:fa2}.

\subsubsection{Weighted hidden-state integration}
Given the sequence of hidden states $\{h_t^{(k)}\}_{k=0}^{K}$ produced by the backbone, we form the final representation via weighted hidden-state integration using the step distribution:
\[
\hat{h}_t=\sum_{k=0}^{K} s_{t,k}\,h_t^{(k)}.
\]
We use $\hat{h}_t$ to produce the logits for next-token prediction. This integration serves two purposes. First, it avoids optimization instability that would arise from discretely choosing a single step count during training. Second, it aligns naturally with token-wise halting at inference: when the step distribution assigns negligible mass to later steps, omitting those steps has little effect on $\hat{h}_t$. As a result, the same learned allocation signal $s$ supports a smooth training objective and a compute-saving inference procedure.

\subsection{Inference}
\label{sec:inference}

At inference time, we perform standard autoregressive decoding and generate one token at a time. For each generation position $t$, the model first computes the step-0 hidden state $h_t^{(0)}$, and the router outputs the step distribution $s_t$, from which we obtain the corresponding mask scores $w_{t,k}$.

We execute pondering steps sequentially starting from $k=1$, and apply token-wise hard stopping based on the mask score. Specifically, we stop as soon as the mask score becomes smaller than a fixed truncation constant $\tau$\footnote{We use a fixed truncation $\tau=10^{-4}$. Under bfloat16 precision, contributions below this magnitude are numerically negligible, so truncating them is effectively equivalent to hard stopping. $\tau$ is not tuned as a halting prior and it is not used as a compute budget hyperparameter.}, and skip all remaining pondering steps for that token. We define the token-specific stopping step
{\setlength{\abovedisplayskip}{6pt}
 \setlength{\belowdisplayskip}{4pt}
 \setlength{\abovedisplayshortskip}{4pt}
 \setlength{\belowdisplayshortskip}{4pt}
\[
\hat{K}_t = \max \{ k \in \{0,\ldots,K\} : w_{t,k} \ge \tau \},
\]
}
and only compute hidden states $\{h_t^{(k)}\}_{k=0}^{\hat{K}_t}$.

We form the integrated representation by truncating the integration at $\hat{K}_t$. During inference, the differentiable attention mask parameterized by $w$ continuously down-weights later-step latent states, while hard stopping avoids executing steps whose remaining probability mass is numerically negligible. As a result, the model executes fewer pondering steps per token, directly reducing inference FLOPs.

\subsection{Training}
\label{sec:training}

\paragraph{Overview.}
Training aims to learn token-wise adaptive pondering end-to-end under purely self-supervised pretraining, while maintaining consistency with the inference-time hard stopping behavior. We perform parallel interleaved updates for efficient training, while adopting Jacobi iterations to align the training dynamics with sequential inference. We also use an auxiliary minimum-ponder penalty to encourage earlier stopping.

\subsubsection{Parallel training via Jacobi iterations}
To align parallel training with sequential inference, we construct an interleaved latent sequence and apply the Transformer iteratively in Jacobi iterations. Let $e(x_t)$ denote the token embedding of the observed token $x_t$, and let $h_t^{(k)}$ be the step-$k$ latent state associated with token position $t$. We form an interleaved sequence (for a given iteration $n$) as
{\setlength{\abovedisplayskip}{3pt}
 \setlength{\belowdisplayskip}{3pt}
 \setlength{\abovedisplayshortskip}{3pt}
 \setlength{\belowdisplayshortskip}{3pt}
\[
\begin{aligned}
S^{(n)} \;=\; \big[\, &e(x_1),\, h_1^{(0)},\, \ldots,\, h_1^{(K-1)},\, e(x_2),\, h_2^{(0)},\, \ldots, \\
&h_2^{(K-1)},\, \ldots \,\big].
\end{aligned}
\]
}

We then apply the same LLM with shared parameters to update all latent states in parallel:
\[
S^{(n+1)} = \mathrm{Transformer}\!\left(S^{(n)}\right),
\]
which produces an updated set of latent states $\{h_t^{(k)}\}$ for all token positions and steps simultaneously.

Crucially, the router is re-applied after each iteration to the updated step-0 states, yielding updated step distributions $s_t$ (and thus updated mask scores). As the iterations proceed, both the latent trajectory and the router outputs are repeatedly refined,
\[
h_t \rightarrow h_t^{\star}, \qquad s_t \rightarrow s_t^{\star},
\]
and after a small number of iterations the process approaches a fixed point. This fixed-point approximation matches the behavior of running pondering steps sequentially at inference, while remaining efficiently parallelizable during training. We provide empirical convergence evidence in Appendix~\ref{app:jacobi_convergence}.

\subsubsection{Auxiliary loss: Minimum-ponder penalty loss}
We optimize the sum of the language modeling loss and an auxiliary penalty:
\[
\mathcal{L} \;=\; \mathcal{L}_{\mathrm{CE}} \;+\; \mathcal{L}_{\mathrm{aux}},
\]
where $\mathcal{L}_{\mathrm{CE}}$ is the standard next-token cross-entropy, and $\mathcal{L}_{\mathrm{aux}}$ encourages earlier stopping when additional pondering steps yield diminishing returns.

To obtain a step-dependent signal of whether early steps already suffice, we define a partial integrated state at truncation depth $i$:
\[
h_t^{(\le i)} \;=\; \sum_{j=0}^{i} s_{t,j}\, h_t^{(j)},
\]
and compute the batch-averaged per-step cross-entropy proxy
\[
ce_i \;=\; \mathbb{E}_{t\in\mathcal{T}}\Big[ \mathrm{CE}\big(\mathrm{LM}(h_t^{(\le i)}),\, x_{t+1}\big) \Big].
\]
We map $ce_i$ to a penalty ratio
\[
\rho_i \;=\; 1 - \sigma\!\left(10\,\big(ce_i - 0.5\big)\right),
\]
so that smaller $ce_i$ yields larger $\rho_i$, more strongly encouraging the model to allocate less probability mass to late steps.

In practice, we instantiate this preference via a ``bottom-mass'' penalty on the mask score values at each step. For each step $k\ge 1$, define
\[
\Delta\rho_k \;=\; \max(\rho_k - \rho_{k-1},\,0),
\]
and penalize the mean of the smallest $\Delta\rho_k$ fraction of the flattened values $w_{\cdot,k}$. The auxiliary loss is
\[
\mathcal{L}_{\mathrm{aux}}
\;=\;
\lambda \sum_{k=1}^{K}
\operatorname{mean}\!\Big(
\operatorname{Bottom}_{\Delta\rho_k}\big(w_{\cdot,k}\big)
\Big),
\]
where $\operatorname{Bottom}_{p}(\cdot)$ denotes the smallest $p$ fraction of elements in a set, and $\lambda$ controls the compute--quality trade-off.

\section{Experiments}
To evaluate token-wise adaptive pondering, we ask whether our approach can turn inference compute from a fixed tax into an allocatable resource while preserving generation quality. We answer this through five targeted experiments: (1) Pareto curves of perplexity versus inference compute under matched pretraining; (2) downstream task performance reported alongside compute; (3) token-difficulty buckets to measure the marginal benefit of additional steps; (4) counterfactual compute shifting (over-prune / under-prune) to localize where compute matters; and (5) ablations over step budget.

\subsection{Pareto Efficiency: Performance vs. Inference Compute}
\label{sec:pareto}

We compare PonderLM-3 against representative baselines that introduce additional computation during autoregressive generation, including PonderLM-1~\cite{zeng2025pretraining1}, PonderLM-2~\cite{zeng2025pretraining2}, and LoopedLM~\cite{giannou2023looped}. We summarize this trade-off with a Pareto curve that compares PPL under matched executed additional computation steps at inference, which serve as a unified proxy for inference FLOPs in our setting.

\paragraph{Settings.}
All models use a 70M-parameter LLaMA-style architecture and are pretrained on a 15B-token subset of The Pile~\cite{gao2020pile} with the same training hyperparameters. For each method, we train three variants with different maximum additional computation steps \(K \in \{2,3,4\}\). At inference, we evaluate each trained variant using its corresponding \(K\), and report perplexity (PPL) together with an inference compute proxy: the average number of executed additional computation steps per token. For fixed-depth baselines, the executed additional steps equal the configured maximum \(K\). For PonderLM-3, they are the actual number of additional computation steps executed after token-wise adaptive pondering skips negligible late steps, and are therefore generally smaller than \(K\).

\begin{figure*}[!htbp] 
  \centering
  \begin{minipage}[c]{0.45\textwidth} 
    \centering
    \includegraphics[width=\textwidth]{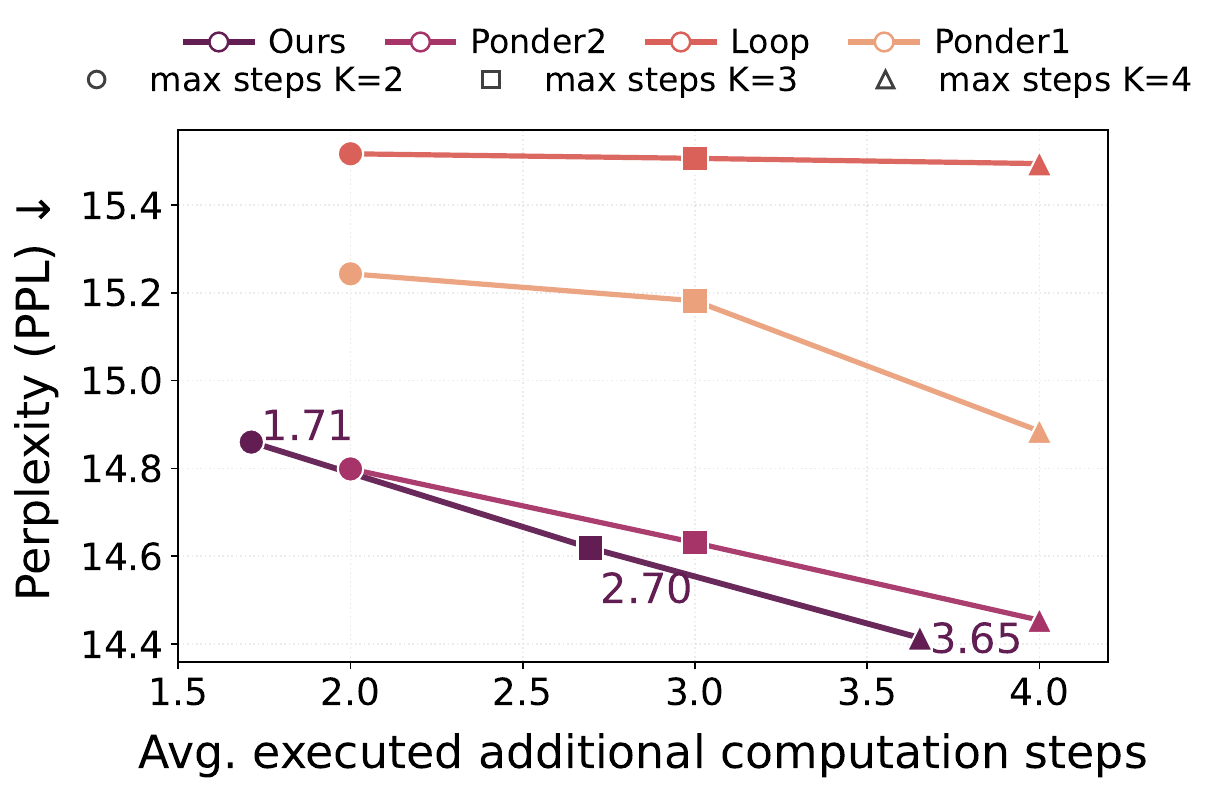}
  \end{minipage}
  \hspace{15 pt}
  \begin{minipage}[c]{0.40\textwidth} 
    \caption{Performance–compute Pareto curve. The x-axis is the average number of executed additional computation steps per token (an inference FLOPs proxy). PonderLM-3 yields a better compute–quality trade-off than the baselines.}
    \label{fig:pareto}
  \end{minipage}
\end{figure*}

\paragraph{Results.}
Figure~\ref{fig:pareto} shows that PonderLM-3 consistently defines a superior Pareto frontier, achieving the lowest perplexity at matched executed compute compared with the fixed-depth and looped baselines. In particular, at the same nominal maximum-step setting \(K\), PonderLM-3 attains comparable or better perplexity while executing fewer additional computation steps on average, and therefore incurs lower inference FLOPs in our setting.

\subsection{Downstream Performance under Compute Budgets}
\label{sec:downstream}

We evaluate the transferability and practical utility of our token-wise adaptive pondering mechanism on a suite of downstream benchmarks.

\paragraph{Settings.}
Following the evaluation protocol in PonderLM-2,  we test on LAMBADA \cite{paperno2016lambada}, SciQ \cite{welbl2017crowdsourcing}, HellaSwag \cite{zellers2019hellaswag}, PIQA\cite{bisk2020piqa}, WinoGrande\cite{sakaguchi2021winogrande}, ARC-Easy / ARC-Challenge \cite{clark2018think}, and RACE\cite{lai2017race}. We include PonderLM-2, PonderLM-1, LoopedLM, MoR~\cite{bae2025mixture} and Pause~\cite{goyal2024thinkspeaktraininglanguage} as baselines under the same downstream evaluation protocol.
We use a 410M-parameter LLaMA-arch model for each baseline, pretrained on a 10B-token subset of The Pile.
Inference compute is reported as estimated inference FLOPs using the standard approximation $\mathrm{FLOPs}\approx 6ND$~\cite{kaplan2020scaling}, where $N$ is the number of parameters and $D$ is the total number of token computations, i.e., the number of generated tokens multiplied by $(1+\overline{s}_{\text{exec}})$ with $\overline{s}_{\text{exec}}$ the average executed additional computation steps per token.


\paragraph{Results.}
Overall, our method remains competitive with strong baselines (especially PonderLM-2) across most tasks, with small gaps on a few datasets, while operating under lower inference compute. Accordingly, Table~\ref{tab:downstream_compute} is not intended to claim universal wins; rather, it shows that our method maintains competitive accuracy under substantially reduced inference compute, yielding a controllable compute--quality trade-off.

\begin{table*}[t]
  \centering
  \small
  \setlength{\tabcolsep}{4.0pt}
  \renewcommand{\arraystretch}{1.12}
  \definecolor{blockgray}{gray}{0.92}

  \caption{Downstream accuracy under reported inference compute (estimated inference FLOPs using $6ND$).}
  \label{tab:downstream_compute}

  \resizebox{\textwidth}{!}{%
  \begin{tabular}{l c @{\hspace{6pt}\vrule\hspace{6pt}} c c c c c c c c c c}
    \toprule
    \textbf{Model} &
    \makecell{\textbf{Inf.\ FLOPs}\\\textbf{(G/token)}} &
    \makecell{Lambada\\OpenAI} &
    \makecell{ARC\\-E} &
    \makecell{Lambada\\Standard} &
    \makecell{ARC\\-C} &
    \makecell{Wino\\Grande} &
    PIQA &
    \makecell{Hella\\Swag} &
    SciQ &
    RACE &
    \textbf{Avg.} \\
    \midrule

    \rowcolor{blockgray}
    \multicolumn{12}{c}{\textbf{\texttt{5-shot}}} \\
    \midrule
    LoopedLM (4 loops)        & 9.84 & 34.2 & 48.2 & 26.6 & 20.6 & 47.0 & 64.2 & 30.7 & 83.4 & 27.8 & 42.5 \\
    Pause (3 pauses)          & 9.84 & 30.7 & 46.9 & 26.3 & 21.8 & 51.6 & 63.4 & 30.1 & 81.8 & 28.5 & 42.3 \\
    MoR (\textit{max} 3 steps)       & 8.44 & 31.5 & 48.5 & 25.3 & 20.0 & 51.0 & 63.7 & 30.6 & 84.7 & 28.0 & 42.6 \\
    PonderLM-1 (3 steps)      & 9.84 & 34.0 & 49.1 & 28.4 & 21.8 & 49.7 & 64.1 & 30.9 & 84.4 & 29.5 & 43.5 \\
    PonderLM-2 (3 steps)      & 9.84 & 37.6 & 50.6 & 34.0 & 21.8 & 53.5 & 65.0 & 32.5 & 87.5 & 31.2 & 46.0 \\
    \textbf{PonderLM-3} (\textit{max} 3 steps)  & \textbf{8.86} & 40.8 & 50.7 & 36.0 & 22.0 & 52.6 & 64.9 & 32.3 & 86.9 & 31.8 & 46.4 \\
    \midrule

    \rowcolor{blockgray}
    \multicolumn{12}{c}{\textbf{\texttt{0-shot}}} \\
    \midrule
    LoopedLM (4 loops)        & 9.84 & 40.4 & 47.2 & 28.6 & 20.4 & 50.8 & 64.4 & 30.6 & 75.4 & 28.8 & 43.0 \\
    Pause (3 pauses)          & 9.84 & 38.1 & 47.5 & 27.8 & 20.3 & 50.3 & 63.8 & 29.8 & 76.3 & 29.0 & 42.5 \\
    MoR (\textit{max} 3 steps)       & 8.44 & 37.0 & 45.3 & 26.6 & 19.9 & 50.1 & 63.8 & 30.0 & 77.3 & 28.6 & 42.0 \\
    PonderLM-1 (3 steps)      & 9.84 & 39.6 & 47.8 & 29.3 & 20.4 & 50.0 & 64.1 & 30.4 & 77.2 & 29.9 & 43.2 \\
    PonderLM-2 (3 steps)      & 9.84 & 45.0 & 47.6 & 36.6 & 21.9 & 51.6 & 64.4 & 32.3 & 80.6 & 31.4 & 45.7 \\
    \textbf{PonderLM-3} (\textit{max} 3 steps)  & \textbf{8.86} & 43.3 & 48.7 & 35.0 & 21.9 & 50.7 & 64.2 & 32.2 & 80.0 & 29.3 & 45.0 \\
    \bottomrule
  \end{tabular}%
  }
  \vskip -0.1in
\end{table*}

\subsection{Where Extra Compute Helps}
\label{sec:analysis-marginal-utility}

In this section, we provide mechanism-level evidence that PonderLM-3 learns to allocate additional computation steps where they matter most. We define a token’s \emph{intrinsic difficulty} as its step-0 token-level prediction error $\ell_t$, i.e., the negative log-likelihood at position $t$ before any additional computation.

\paragraph{Settings.}
We run this analysis with a 410M-parameter PonderLM-3 model pretrained on a 10B-token subset of The Pile with maximum additional computation steps $K{=}3$. Unless otherwise specified, statistics are computed on a held-out Pile validation slice comprising on the order of $10^7$ tokens, and subsequent analyses reuse the same setting. 

\paragraph{Difficulty buckets and marginal-utility metric.}
We partition tokens into easy, medium, and hard buckets based on $\ell_t$. Let $CE_i$ denote token-level cross-entropy after executing $i$ additional computation steps. We define the marginal utility of step $i$ as
\[
\Delta CE_i \triangleq CE_{i-1} - CE_i,
\]
and aggregate $\Delta CE_i$ within each bucket.

\begin{figure*}[!htbp]
  \centering
  \captionsetup[subfigure]{justification=centering}
  
  \begin{subfigure}[b]{0.45\textwidth}
    \centering
    \includegraphics[width=\linewidth]{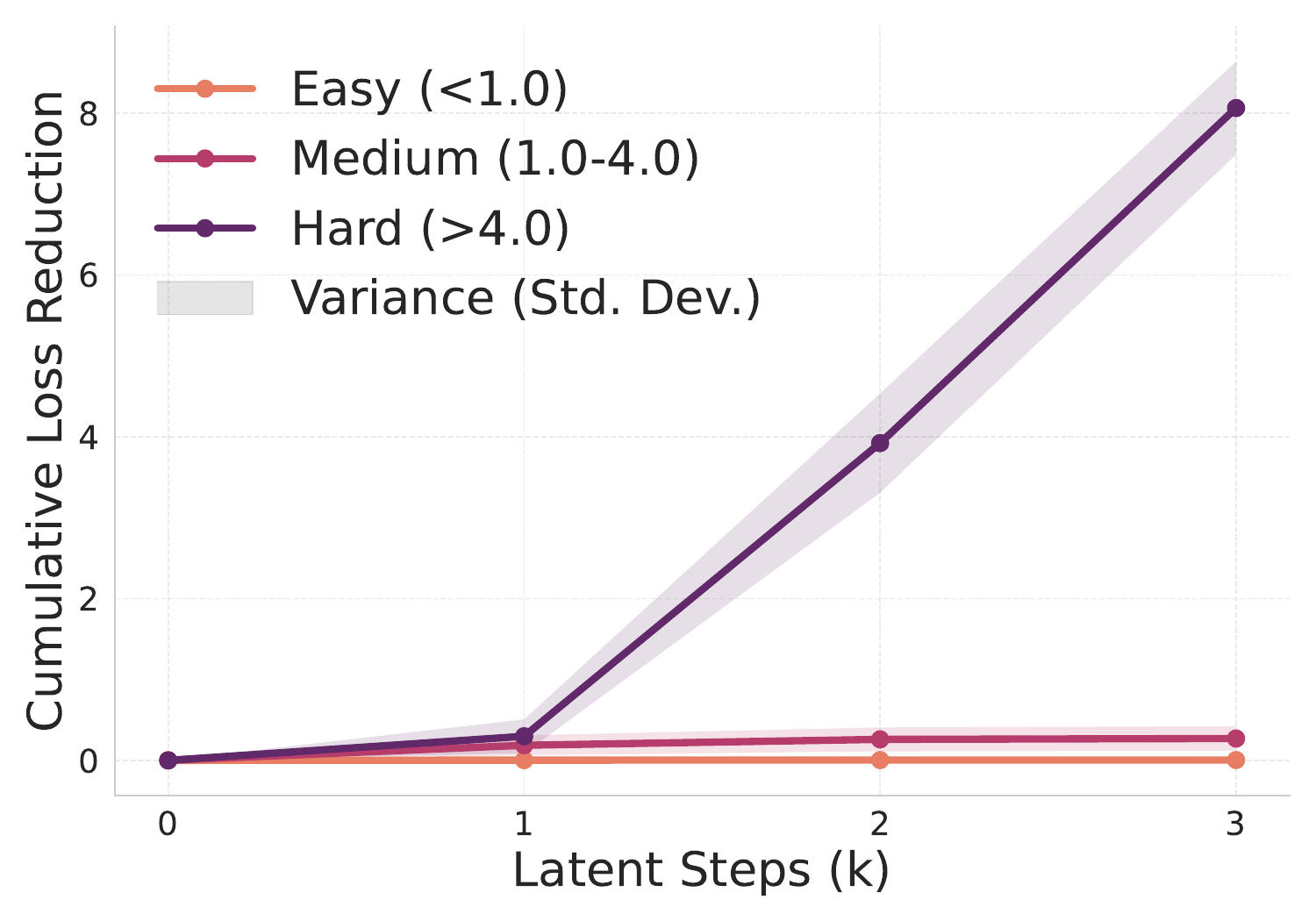}
    \caption{}
  \end{subfigure}
  \hspace{0.05\textwidth}
  \begin{subfigure}[b]{0.45\textwidth}
    \centering
    \includegraphics[width=\linewidth]{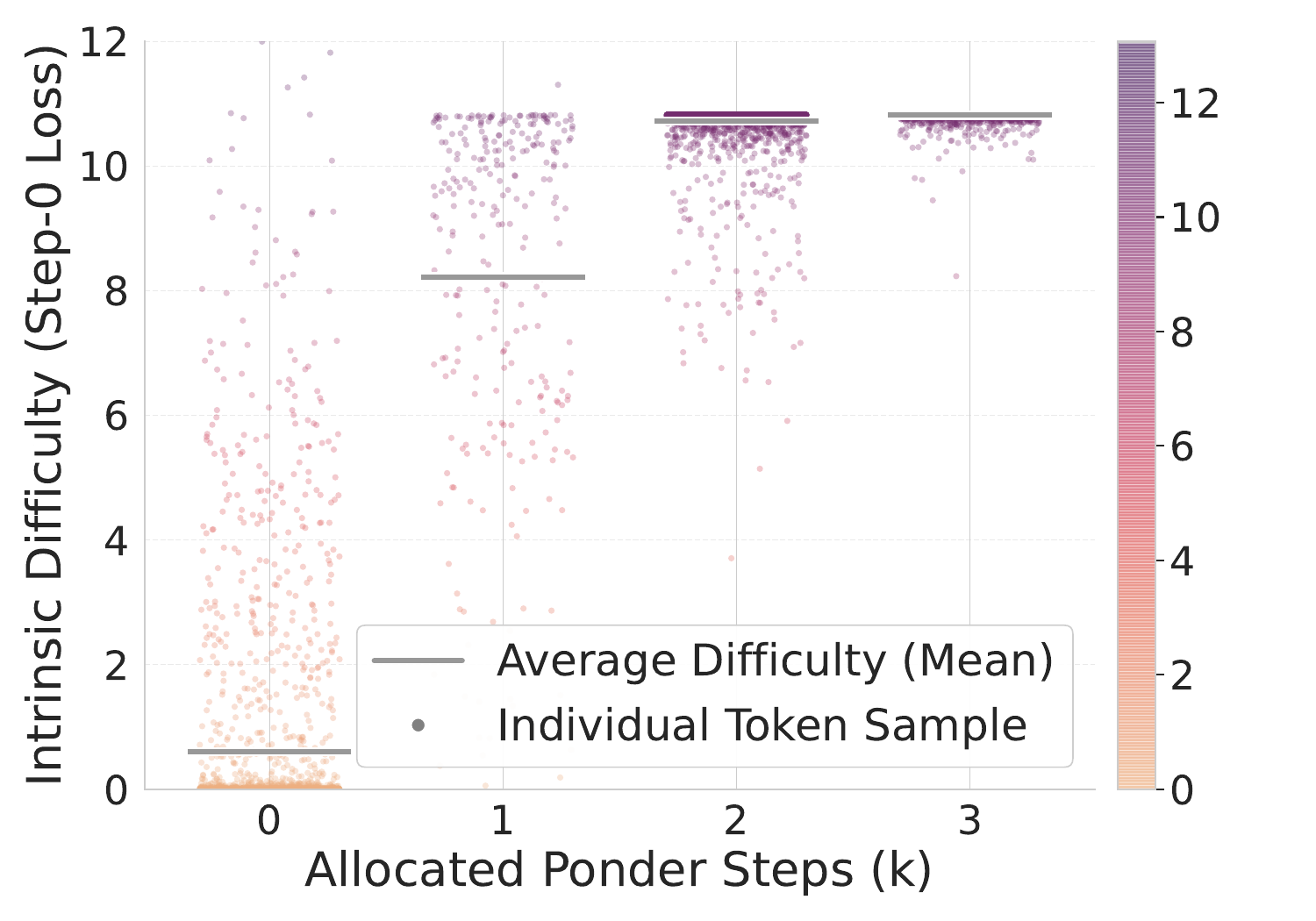}
    \caption{}
  \end{subfigure}
  
  \caption{Where extra compute helps. (a) Bucket-wise $\Delta CE_i$ over executed additional computation steps. (b) Intrinsic difficulty $\ell_t$ versus executed additional computation steps.}
  \label{fig:difficulty_utility}
\end{figure*}

\paragraph{Difficulty-aware utility and allocation.}
Figure~\ref{fig:difficulty_utility}(a) shows that extra steps deliver substantially larger gains on hard tokens, while improvements on easy tokens saturate quickly. Figure~\ref{fig:difficulty_utility}(b) shows that tokens receiving more executed additional computation steps have higher intrinsic difficulty $\ell_t$. Taken together, these results indicate that PonderLM-3 assigns compute where it yields the largest returns---spending more steps on hard tokens and avoiding unnecessary steps on easy tokens---which helps explain the favorable inference compute--quality trade-offs observed earlier.

We next perform a counterfactual stress test of the learned router: we deliberately shift the step distribution at inference time to either remove compute (over-prune) or add compute (under-prune), and examine where the resulting loss changes occur. Under a meaningful allocation, taking compute away should hurt primarily the tokens that benefit from it, whereas adding compute to low-return tokens should produce only limited gains.

\begin{figure}[!htbp]
  \centering
  
  \begin{minipage}[b]{0.56\textwidth}
    \centering
    \includegraphics[width=\linewidth]{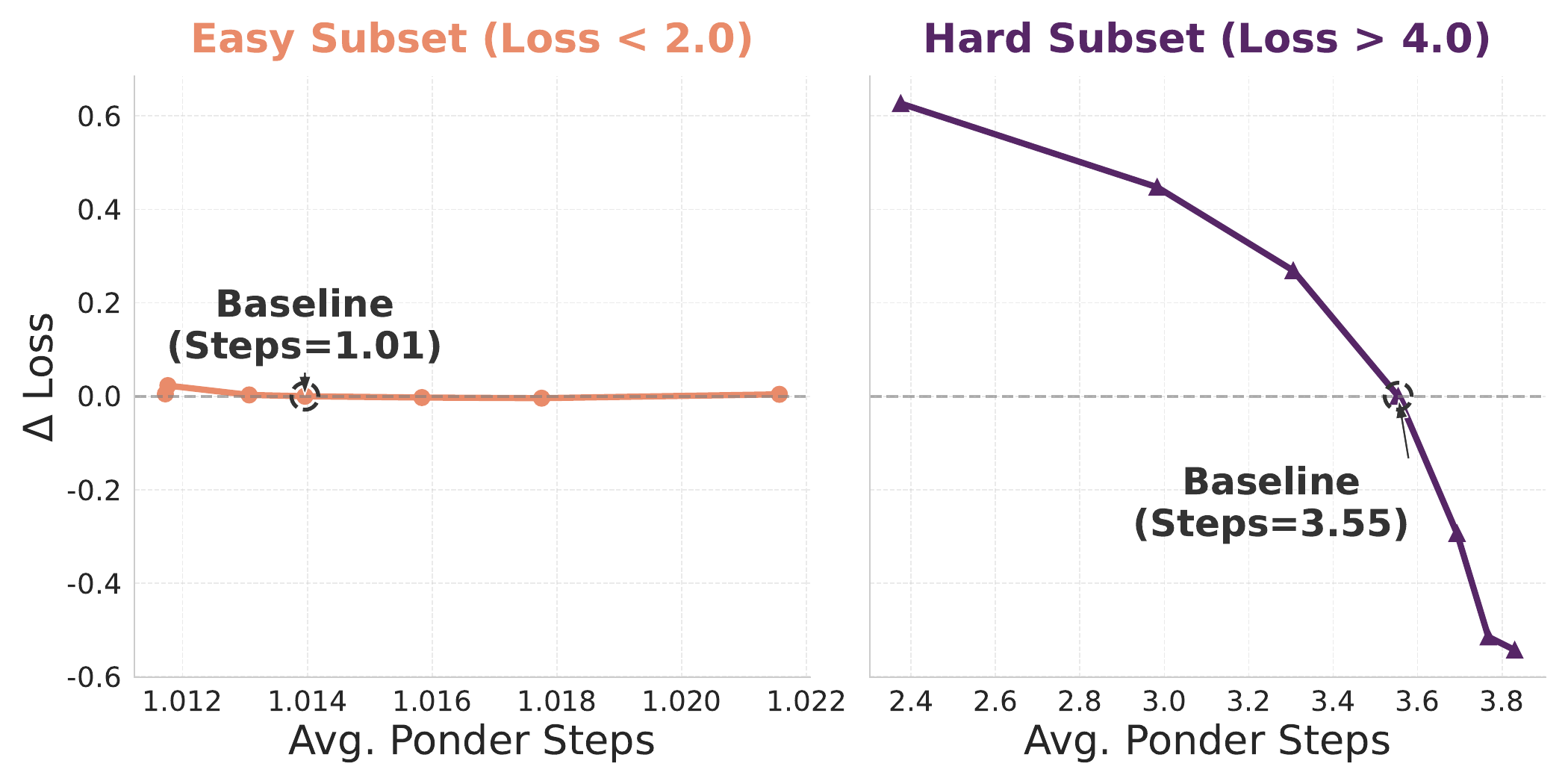}
    \caption{Counterfactual compute shifting by sweeping the inference-time router logit bias $\alpha$. We report $\Delta$Loss relative to the adaptive baseline ($\alpha=0$) versus the induced average executed additional computation steps (left: easy subset; right: hard subset).}
    \label{fig:counterfactual_shift}
  \end{minipage}
  \hfill 
  \begin{minipage}[b]{0.40\textwidth}
    \centering
    \includegraphics[width=\linewidth]{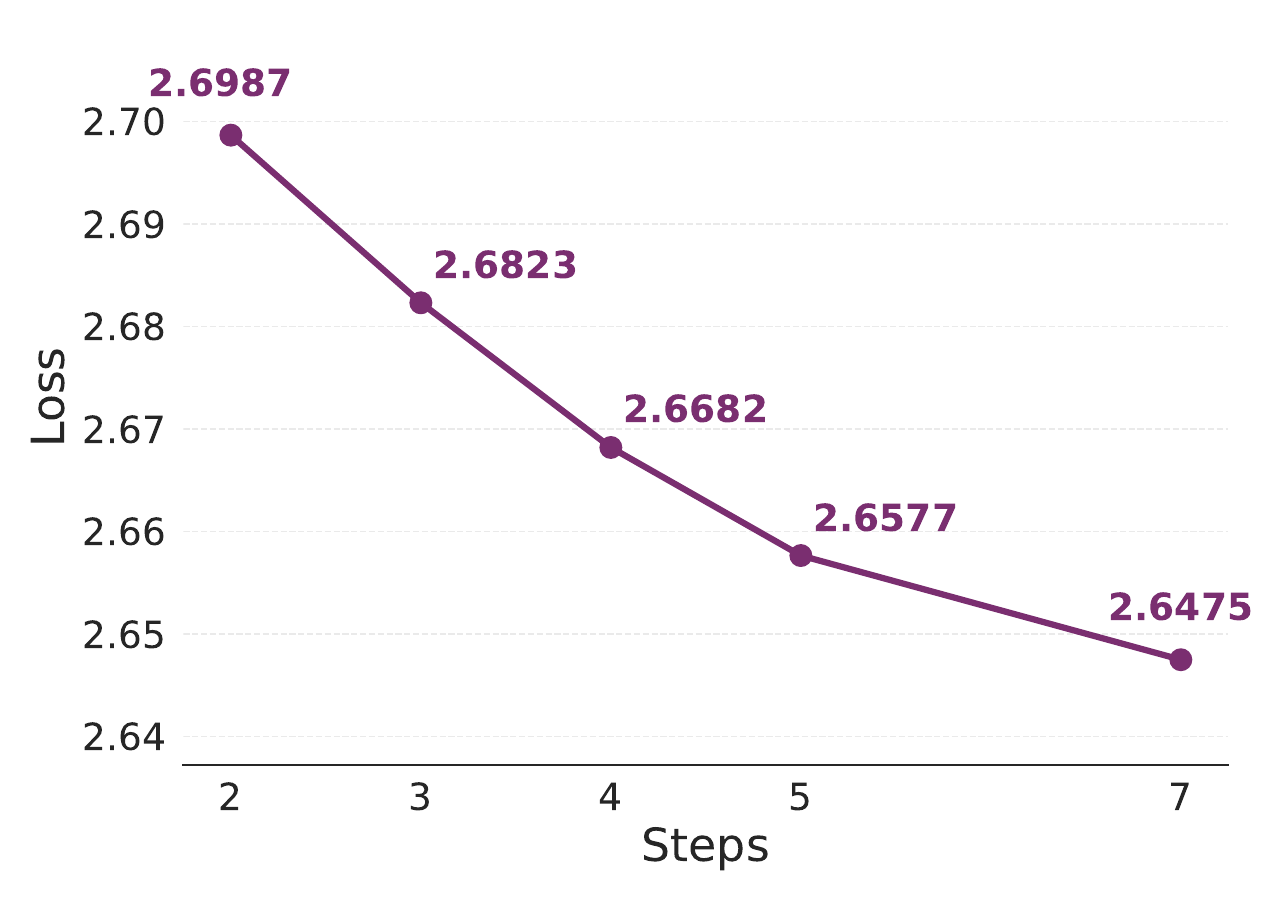}
    \caption{Scaling analysis of performance versus depth. Pretraining cross-entropy loss decreases as the maximum number of additional computation steps $K$ increases.}
    \label{fig:scaling-depth}
  \end{minipage}
  
  \vskip -0.1in
\end{figure}

\paragraph{Intervention: biasing the router at inference time.}
To smoothly control inference-time compute without changing any trained weights, we intervene on the router by adding a tunable bias $\alpha$ directly to its pre-softmax logits. Specifically, for each token $t$, we modify the step distribution as
\[
s_t=\mathrm{Softmax}\!\left(W_{\text{ponder}}\,h_t^{(0)}+\alpha\right),
\]
where $\alpha$ is applied only at inference. Negative $\alpha$ shifts probability mass toward smaller step counts, leading to more aggressive pruning (over-prune), while positive $\alpha$ shifts mass toward larger step counts, encouraging deeper computation (under-prune). We then derive the corresponding mask score $w_{t,k}$ from the biased $s_t$ and run inference with the same stopping rule. We report compute on the x-axis in Figure~\ref{fig:counterfactual_shift} as the induced pondering steps per token.

\paragraph{Results: compute shifts primarily affect hard tokens.}
Figure~\ref{fig:counterfactual_shift} reports the loss change $\Delta$Loss relative to the adaptive baseline ($\alpha=0$) for easy and hard subsets. The easy subset is largely insensitive across a wide range of compute shifts, indicating diminishing returns from spending additional computation on easy tokens. In contrast, the hard subset is highly sensitive: over-pruning increases loss sharply, whereas under-pruning yields noticeable improvements. Together, these results show that PonderLM-3’s token-wise adaptive pondering learns a non-uniform sensitivity to compute, making inference performance robust to pruning on easy tokens while preserving the ability to benefit from extra compute on hard ones.

\subsection{Ablations and Design Choices}
\label{sec:ablation}

We ablate a key design choice that affects both generation quality and inference compute: the maximum number of additional computation steps $K$. Since our goal is to understand the compute--quality trade-off induced by the maximum pondering depth, we report pretraining cross-entropy loss as a function of the hyperparameter $K$ (the maximum number of pondering steps).

We train 70M-parameter LLaMA-style models on a 15B-token subset of The Pile and sweep the maximum number of pondering steps $K \in \{2,3,4,5\}$. Figure~\ref{fig:scaling-depth} shows that increasing $K$ consistently lowers pretraining cross-entropy in our setting, suggesting that additional pondering steps provide extra capacity for refining token representations. In practice, larger $K$ also increases training and evaluation cost. Considering the resource budget and diminishing returns at higher depths, we choose $K=3$ as the default setting in all main results.

\section{Conclusion}
\label{sec:conclusion}

We introduced PonderLM-3, which enables token-wise adaptive pondering to allocate additional computation on a per-token basis, while remaining end-to-end differentiable through a differentiable attention mask. Across pretraining evaluations, PonderLM-3 achieves better performance than PonderLM-1, PonderLM-2, LoopedLM, and Pause models at matched inference FLOPs, yielding a steeper and more favorable perplexity--compute Pareto frontier. On downstream benchmarks, PonderLM-3 attains performance comparable to fixed-step PonderLM-2 trained with the same maximum number of pondering steps, while using fewer inference FLOPs in practice. Overall, PonderLM-3 provides a simple and effective framework for making inference compute a controllable, token-adaptive resource in pretrained language models.

\bibliography{references}

\newpage
\appendix
\section{FlashAttention-2 Compatibility for Differentiable Attention Mask}
\label{appendix:fa2}
FlashAttention-2 (FA2) kernels typically support boolean attention masks (causal and padding) but do not natively expose an interface for adding a dense, real-valued bias term to the attention logits.


\begin{figure}[!htbp]
\centering
\scriptsize
\setlength{\tabcolsep}{4pt}
\renewcommand{\arraystretch}{1.2}
\begin{tabular}{l|ccccccccc}
\toprule
 & $x_1$ & $h_1^{(0)}$ & $h_1^{(1)}$ & $x_2$ & $h_2^{(0)}$ & $h_2^{(1)}$ & $x_3$ & $h_3^{(0)}$ & $h_3^{(1)}$ \\
\midrule
$x_1$         & $1$ &     &     &     &     &     &     &     &     \\
$h_1^{(0)}$   & $1$ & $w_{1,0}$ &     &     &     &     &     &     &     \\
$h_1^{(1)}$   & $1$ & $w_{1,0}$ & $w_{1,1}$ &     &     &     &     &     &     \\
\midrule
$x_2$         & $1$ & $w_{1,0}$ & $w_{1,1}$ & $1$ &     &     &     &     &     \\
$h_2^{(0)}$   & $1$ & $w_{1,0}$ & $w_{1,1}$ & $1$ & $w_{2,0}$ &     &     &     &     \\
$h_2^{(1)}$   & $1$ & $w_{1,0}$ & $w_{1,1}$ & $1$ & $w_{2,0}$ & $w_{2,1}$ &     &     &     \\
\midrule
$x_3$         & $1$ & $w_{1,0}$ & $w_{1,1}$ & $1$ & $w_{2,0}$ & $w_{2,1}$ & $1$ &     &     \\
$h_3^{(0)}$   & $1$ & $w_{1,0}$ & $w_{1,1}$ & $1$ & $w_{2,0}$ & $w_{2,1}$ & $1$ & $w_{3,0}$ &     \\
$h_3^{(1)}$   & $1$ & $w_{1,0}$ & $w_{1,1}$ & $1$ & $w_{2,0}$ & $w_{2,1}$ & $1$ & $w_{3,0}$ & $w_{3,1}$ \\
\bottomrule
\end{tabular}
\caption{Soft attention mask over an interleaved sequence. Rows correspond to \textit{queries}, and columns correspond to \textit{keys}. Entries show the multiplicative mask values applied to attention weights. Blank entries are masked out by causality or padding.}
\label{fig:soft-mask-table}
\end{figure}

\paragraph{Soft attention mask pattern.}
Figure~\ref{fig:soft-mask-table} visualizes the differentiable attention mask used in our interleaved sequence of observed tokens and latent states. In our formulation, the attention logits contain two masking components: (i) the standard boolean mask $M$ (causal and padding), and (ii) an additional dense bias matrix $G$ induced by the mask score $w$. While $M$ is supported by FA2, the dense, real-valued bias $G$ is not. We therefore construct an equivalent computation that preserves the FA2 interface by modifying $Q$, $K$, and $V$ (and correspondingly the head dimension) so that adding $G$ is realized implicitly through a standard dot-product attention call.

\paragraph{Target attention form.}
The original attention is
\[
\mathrm{Softmax}\!\left(\frac{QK^\top}{\sqrt{d}} + M\right)V,
\]
and we aim to compute
\[
\mathrm{Softmax}\!\left(\frac{QK^\top}{\sqrt{d}} + M + G\right)V.
\]
Here
\[
G=\log(\text{ponder mask score})=\log w,
\]
and $G$ takes the following broadcasted matrix form:
\[
G=
\begin{bmatrix}
\log w_1 & \log w_2 & \cdots & \log w_n\\
\log w_1 & \log w_2 & \cdots & \log w_n\\
\vdots   & \vdots   & \ddots & \vdots\\
\log w_1 & \log w_2 & \cdots & \log w_n
\end{bmatrix}\in\mathbb{R}^{n\times n}.
\]
That is, the additive bias is constant across query rows and depends only on the key position, matching the soft attention pattern in Figure~\ref{fig:soft-mask-table}.

\paragraph{FA2-compatible equivalent construction.}
Since FA2 does not directly support adding $G$ to attention logits, we realize the same effect by augmenting the attention computation such that
\[
\frac{Q' {K'}^\top}{\sqrt{d'}} \;=\; \frac{QK^\top}{\sqrt{d}} + G.
\]
We augment the head dimension by one:
\[
Q\in\mathbb{R}^{n\times d},\quad K\in\mathbb{R}^{n\times d},\quad d' = d+1,
\]
and define
\[
Q'=\big[\,Q\;\; q_{\text{extra}}\,\big]\in\mathbb{R}^{n\times d'},
\qquad
K'=\big[\,K\;\; k_{\text{extra}}\,\big]\in\mathbb{R}^{n\times d'}.
\]
Then
\[
\frac{Q'{K'}^\top}{\sqrt{d'}}
=
\frac{\big[\,Q\;\; q_{\text{extra}}\,\big]
\begin{bmatrix}
K^\top\\
k_{\text{extra}}^\top
\end{bmatrix}}{\sqrt{d'}}
=
\frac{QK^\top + q_{\text{extra}}k_{\text{extra}}^\top}{\sqrt{d'}}
=
\frac{QK^\top}{\sqrt{d}} + G.
\]
Therefore it suffices to enforce
\[
q_{\text{extra}}k_{\text{extra}}^\top \;=\; \sqrt{d'}\cdot G.
\]
A concrete choice is
\[
q_{\text{extra}}=
\begin{bmatrix}
\sqrt{d'}\\
\sqrt{d'}\\
\vdots\\
\sqrt{d'}
\end{bmatrix}
\in\mathbb{R}^{n\times 1},
\qquad
k_{\text{extra}}=
\begin{bmatrix}
\log w_1\\
\log w_2\\
\vdots\\
\log w_n
\end{bmatrix}
\in\mathbb{R}^{n\times 1},
\]
which yields
\[
q_{\text{extra}}k_{\text{extra}}^\top
=
\sqrt{d'}
\begin{bmatrix}
\log w_1 & \log w_2 & \cdots & \log w_n\\
\log w_1 & \log w_2 & \cdots & \log w_n\\
\vdots   & \vdots   & \ddots & \vdots\\
\log w_1 & \log w_2 & \cdots & \log w_n
\end{bmatrix}
=
\sqrt{d'}\cdot G,
\]
and thus implements the desired dense bias through a standard scaled dot product in the augmented space.

\paragraph{Augmenting values.}
Because the augmented attention operates in dimension $d'$, we also augment values:
\[
V'=\big[\,V\;\; v_{\text{extra}}\,\big]\in\mathbb{R}^{n\times d'},
\]
with
\[
v_{\text{extra}}=\mathbf{0}\in\mathbb{R}^{n\times 1}.
\]
This ensures the additional channel contributes no content to the output, and the resulting attention call can be executed using the FA2 interface with the original boolean mask $M$, while being equivalent to adding the dense bias $G$ to the logits.

\section{Convergence of Jacobi-Style Parallel Iterations}
\label{app:jacobi_convergence}

This appendix provides empirical evidence that the Jacobi-style parallel iterations used during training rapidly approach the fixed-point representations induced by sequential inference. The key point is that, at each training iteration, the latent states and the router outputs are \emph{synchronously} refined, yielding a stable trajectory that quickly aligns with inference-time behavior.

\paragraph{Synchronous updates of latent states and routing signals.}
At iteration $n$, we form an interleaved sequence
\[
S^{(n)}=\big[e(x_1),\,h_1^{(0)},\,\ldots,\,h_1^{(K-1)},\,e(x_2),\,h_2^{(0)},\,\ldots,\,h_2^{(K-1)},\,\ldots\big],
\]
and apply a Transformer with shared parameters to update all latent states in parallel:
\[
S^{(n+1)}=\mathrm{Transformer}\!\left(S^{(n)}\right).
\]
This produces an updated set of latent states $\{h_t^{(k)}\}$ for all token positions and steps simultaneously. Crucially, after each iteration we re-apply the router to the updated step-0 states, yielding updated step distributions $s_t$ (and thus updated mask scores). As the iterations proceed, both the latent trajectory and the router outputs are repeatedly refined,
\[
h_t \rightarrow h_t^{*}, 
\qquad 
s_t \rightarrow s_t^{*},
\]
and after a small number of iterations the process approaches a fixed point. This fixed-point approximation matches the behavior of running computation steps sequentially at inference, while remaining efficiently parallelizable during training.

\begin{figure}[t]
    \centering
    \includegraphics[width=0.6\linewidth]{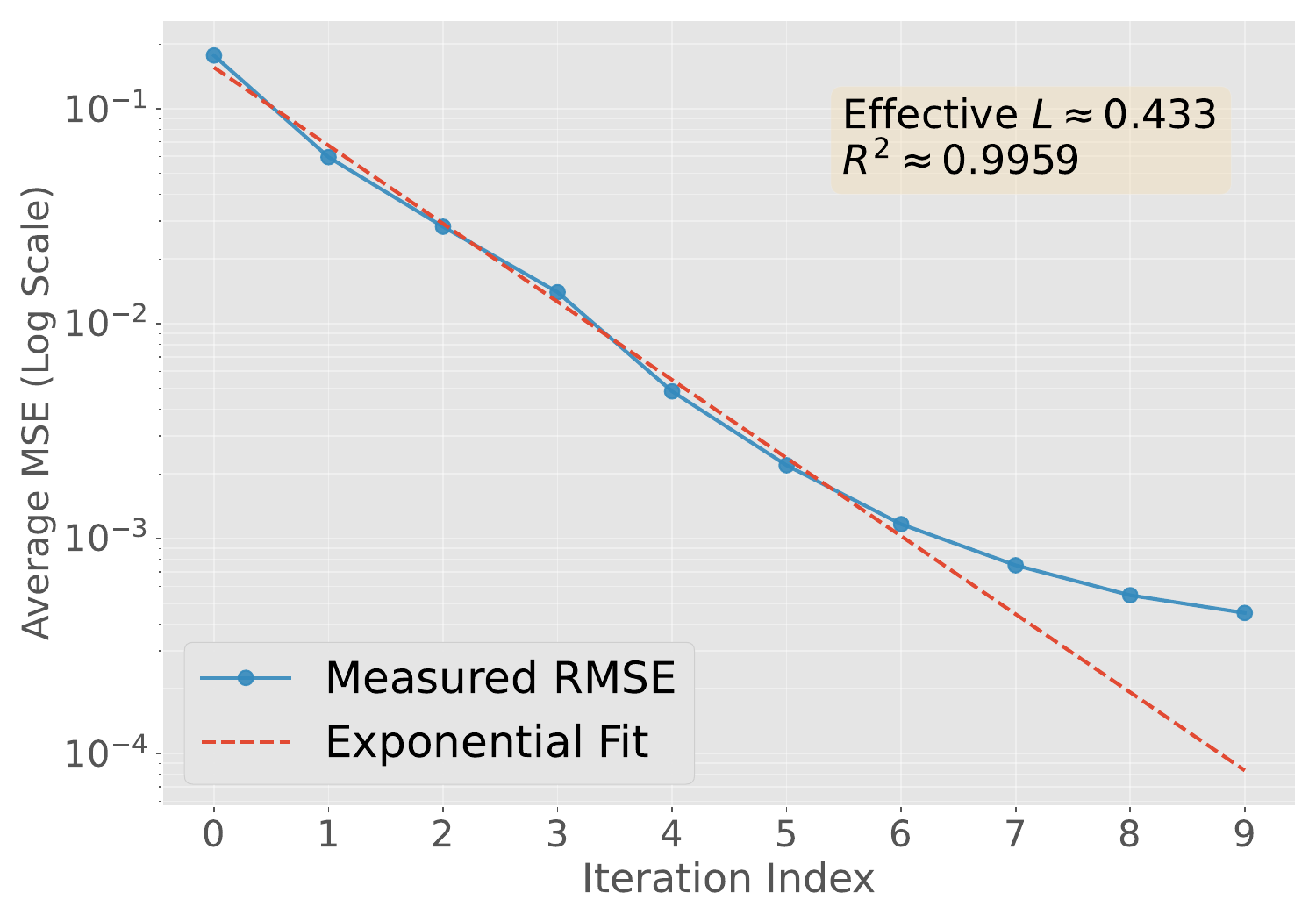}
    \caption{RMSE convergence of Jacobi-style parallel iterations.
    Average RMSE between training-iteration latent states $h^{(n)}$ and the inference-time fixed point $h^{*}$ over iteration index $n$ (log scale). The curve decays rapidly and is well approximated by an exponential fit (effective $L \approx 0.433$, $R^{2}\approx 0.9959$), indicating fast approach to the inference-time solution.}
    \label{fig:rmse_convergence}
\end{figure}

\paragraph{RMSE-to-fixed-point convergence.}
To quantify how quickly the training-time iterations approach inference-time behavior, we measure the discrepancy between the latent states produced after $n$ training iterations and the corresponding inference-time fixed-point states. Concretely, we compute the root-mean-square error (RMSE) between $h^{(n)}$ and the inference-time fixed-point $h^{*}$, and track the average RMSE over iterations. Figure~\ref{fig:rmse_convergence} shows that the RMSE decreases rapidly over the first few iterations and then levels off, indicating that the parallel iterations quickly approach the inference-time solution.

We additionally fit an exponential decay model to the measured RMSE curve to characterize the convergence rate. The fit yields an effective contraction factor $L \approx 0.433$ with $R^{2}\approx 0.9959$, suggesting near-exponential convergence: on average, each additional iteration reduces the RMSE by a constant multiplicative factor ($L<1$), and the high $R^2$ indicates that this trend explains the observed decay well. Together, these results support that the Jacobi-style parallel refinement provides a fast approximation to the inference-time fixed point, thereby justifying the train--inference alignment used throughout the paper.

\section{Analysis of Pondering Step Distribution}

To elucidate the underlying logic of the learned halting policy, we conduct a quantitative analysis of the pondering step distribution across distinct linguistic contexts.

\subsection*{Empirical Evidence of Difficulty-Aware Allocation}

Tables \ref{tab:token_stats} and \ref{tab:compound_stats} reveal the statistical divergence in computational depth, highlighting how the model dynamically modulates pondering depth based on predictive uncertainty. We observe two dominant patterns in the emergent halting behavior:

\begin{enumerate}
    \item \textbf{Anticipatory Halting in Functional Tokens:} As shown in Table \ref{tab:token_stats}, deterministic functional tokens such as ``the'' and ``of'' frequently trigger early-exit mechanisms. In stark contrast, the indefinite article ``a'' almost exclusively demands maximal depth (Avg Step $\approx$ 3.0). 
    \item \textbf{Boundary-Aware Computation in Compound Words:} Table \ref{tab:compound_stats} illustrates a striking bifurcation in halting behavior within compound words. For \textbf{non-terminal tokens}, the model frequently triggers early halting (with nearly 30\% exiting before the maximum depth). However, for \textbf{terminal tokens}—which must anticipate the transition to a new semantic unit—the policy allocates maximal computation (\textbf{100\% Category 3}) without exception.
\end{enumerate}

\begin{table}[h]
\centering
\begin{minipage}{0.45\textwidth}
    \centering
    \caption{Pondering step statistics for individual tokens, highlighting the distinction between high-uncertainty articles and deterministic particles.}
    \label{tab:token_stats}
    \vspace{2mm}
    \begin{tabular}{lccc}
    \toprule
    \textbf{Token} & \textbf{Count} & \textbf{Avg Step} & \textbf{Pruned \%} \\
    \midrule
    \textbf{a}    & 4,203  & \textbf{2.999} & 0.07\% \\
    \textbf{the} & 14,727 & 2.257 & 74.28\% \\
    \textbf{of}  & 6,840  & 2.337 & 66.27\% \\
    \bottomrule
    \end{tabular}
\end{minipage}
\hfill
\begin{minipage}{0.52\textwidth}
    \centering
    \caption{Halting distribution in Compound Words. The policy prioritizes computation for word boundaries over highly predictable intra-word completions.}
    \label{tab:compound_stats}
    \vspace{2mm}
    \begin{tabular}{lrr}
    \toprule
    \textbf{Category} & \textbf{Non-terminal} & \textbf{Terminal} \\
    \midrule
    Category 0 & 6.24\%  & 0.00\% \\
    Category 1 & 8.13\%  & 0.00\% \\
    Category 2 & 13.96\% & 0.00\% \\
    Category 3 & \textbf{71.67\%} & \textbf{100.00\%} \\
    \midrule
    \textbf{Total Count} & 5,980 & 2,164 \\
    \bottomrule
    \end{tabular}
\end{minipage}
\end{table}

\subsection*{Discussion: Principled Allocation as a Foundation for Test-Time Scaling}

The empirical evidence presented in Tables \ref{tab:token_stats} and \ref{tab:compound_stats} collectively validates a core mechanistic claim: the model does not allocate compute based on static word features, but rather through an \textbf{active sensitivity to predictive entropy}. The internal logic follows a consistent hierarchical structure:

First, the model demonstrates a precise understanding of \textbf{syntactic uncertainty}. The sharp divergence between ``a'' and ``the'' reveals that the halting policy is sensitive to the combinatorial possibilities following a token. While both are high-frequency particles, the model correctly identifies the indefinite article as a "high-entropy" junction that requires deeper computation to prepare for a diverse range of potential noun phrases. 

Second, this logic extends to \textbf{semantic boundaries} within compound words. The bifurcation between intra-word and boundary tokens proves the model can distinguish between "trivial" local completions and "high-stakes" semantic transitions. By triggering early-exit for predictable internal tokens while sustaining maximal depth at word boundaries, the policy effectively exploits local determinism to save budget for critical information bottlenecks.

Ultimately, these combined patterns confirm that the learned mechanism provides the necessary \textbf{functional foundation for Test-time Scaling}. Effective scaling requires the model to prioritize its "pondering" budget where it yields the highest marginal utility. By accurately concentrating compute on high-entropy positions—be they indefinite syntactic junctions or semantic word boundaries—our mechanism ensures that additional inference-time computation is translated into genuine generation quality, rather than being wasted on deterministic transitions.
\end{document}

%% file: notations.tex
\usepackage{xspace}

\theoremstyle{plain}

\newtheorem*{proposition*}{Proposition}

\theoremstyle{definition}

\theoremstyle{definition}

\def\eqref#1{equation~\ref{#1}}

